%% file: main.tex
\documentclass{article}
\usepackage{spconf,amsmath,graphicx}
\usepackage{hyperref}
\usepackage{mathtools}
\usepackage{tcolorbox}
\usepackage{color}
\usepackage{theorem}
\usepackage{times,amsmath,epsfig}
\usepackage{amssymb}
\usepackage{relsize}
\usepackage{subfigure}
\usepackage{cite}
\usepackage{cases}
\usepackage{siunitx}
\usepackage{algorithmic}
\usepackage{algorithm}
\usepackage{lipsum} 
\input{my_sections.tex}
\usepackage{tikz}
\usetikzlibrary{shapes,arrows}
\input{mysymbol.sty}
\input{figures/tikz_styles}

\newtheorem{myproposition}{\bf Proposition}
\newtheorem{problem}{\bf Problem}

\title{Label Propagation across Graphs: Node Classification using Graph Neural Tangent Kernels}

\name{Artun Bayer, Arindam Chowdhury, and Santiago Segarra
\thanks{Emails:  \{ab116, ac131, segarra\}@rice.edu. }}
\address{Electrical and Computer Engineering, Rice University, USA}
\begin{document}
\ninept
\maketitle
\begin{abstract}
    Graph neural networks (GNNs) have achieved superior performance on node classification tasks in the last few years. 
    Commonly, this is framed in a transductive semi-supervised learning setup wherein the entire graph -- including the target nodes to be labeled -- is available for training. 
    Driven in part by scalability, recent works have focused on the inductive case where only the labeled portion of a graph is available for training. 
    In this context, our current work considers a challenging inductive setting where a set of labeled graphs are available for training while the unlabeled target graph is completely separate, i.e., there are no connections between labeled and unlabeled nodes.
    Under the implicit assumption that the testing and training graphs come from similar distributions, our goal is to develop a labeling function that generalizes to unobserved connectivity structures. 
    To that end, we employ a graph neural tangent kernel (GNTK) that corresponds to infinitely wide GNNs to find correspondences between nodes in different graphs based on both the topology and the node features. 
    We augment the capabilities of the GNTK with residual connections and empirically illustrate its performance gains on standard benchmarks.
\end{abstract}
%
\section{Introduction}\label{sec:intro}

In this era of information revolution, data processing systems and methods are tasked with generating suitable representations of enormous amounts of data obtained from multiple domains and modalities. 
In addition to the individual attributes and features, it is crucial to extract essential relationships among multiple data points. 
Graphs have become ubiquitous in modern data processing and algorithms on account of the versatility that they offer in representing relational structures~\cite{shuman2013emerging,huang2014big}. 
The availability of vast quantities of network data have also paved the way for application of advanced signal processing~\cite{ortega2018graph,mateos2019connecting} and machine learning tools~\cite{wu2020comprehensive,zhang2020deep} for graph representation learning.

Among these generalizations, the most prominent examples are graph neural networks (GNNs). 
A variety of GNN architectures~\cite{kipf2016semi,hamilton2017inductive,chen2018fastgcn}, both centralized and distributed~\cite{wang2021flexgraph,wolfe2021gist} have been successfully applied to solve challenging problems in multiple fields. 
While these architectures vary in structure and scope, their core operating principle involves combining the information stored in connected nodes to generate useful node-level and graph-level embeddings~\cite{hamilton2017inductive} that can be leveraged for multiple downstream tasks like classification, regression, and clustering~\cite{chen2020graph}. 
In particular, node classification~\cite{hu2019hierarchical,oono2019graph} is a well-studied application of node representation learning. 

Gradient descent (GD)~\cite{ruder2016overview} is, arguably, the most popular learning algorithm for these connectionist methods. 
Its basic operation involves taking a sequence of gradient steps towards a desired optimum. 
However, tuning its operating conditions can be quite tricky~\cite{bottou2012stochastic}. 
Current research has shown that under certain limiting conditions on the neural architectures, GD resembles kernel regression with a specialized deterministic kernel, namely, the Neural Tangent Kernel (NTK)~\cite{jacot2018neural}. 
Its utility lies in eliminating the necessity of explicit gradient updates for training. Closed-form expressions to determine the NTK corresponding to multi-layer perceptrons (MLPs)~\cite{jacot2018neural} and convolutional neural networks (CNNs)~\cite{arora2019exact} have been determined. 
More recently, these methods have been extended to GNNs facilitating their fusion with the more classical graph kernel approach by deriving a specialized kernel, called Graph Neural Tangent Kernel (GNTK)~\cite{du2019graph}, from an infinitely wide GNN.    

The GNTK has been primarily applied to determine similarities between graphs, which are then used for graph classification tasks~\cite{du2019graph}. 
In this work, we focus on a different capability of the GNTK and leverage it to find similarities among nodes in a graph, which can be used for node classification. 
Specifically, our approach caters to a purely inductive setting~\cite{hamilton2017inductive} wherein the node labels of a previously-unseen and completely-unlabeled target graph are to be predicted based on a set of training graphs with fully-labeled nodes. 
This is more challenging than the typical transductive setting wherein the target graph is available during training, albeit without any labels~\cite{kipf2016semi}. 
A formulation of this form is useful for ego-networks~\cite{you2021identity} and wireless networks~\cite{chowdhury2021unfolding} among several other scenarios, where the aim is to extract representations of a new instance based on a labeled set of identically sampled instances.
Further, inspired by the general observation that residual connections tend to improve overall performance of GNNs~\cite{xu2021optimization}, we have established a framework to extract GNTKs corresponding to GNNs with residual connections. 
Our empirical analysis suggests that there is a clear performance gain due to the proposed augmentation.    

\medskip\noindent {\bf Contributions.}
The contributions of this paper are threefold:\\
1) We provide a closed-form recurrent expression for the computation of a GNTK associated with a GNN with residual connections.\\
2) We incorporate the derived GNTK into a node classification pipeline to solve the inductive problem of estimating the labels of a completely unlabeled graph.\\
3) Through numerical experiments, we illustrate the benefit of incorporating residual connections into the GNTK and compare the performance of the proposed node classification pipeline against GNN baselines.

\section{Preliminaries and Problem formulation}\label{S:Formulation}

In Section~\ref{Ss:graphs_and_gnns} we introduce the concepts of graphs and GNNs and in Section~\ref{Ss:ntks} we present NTKs.
We provide a precise formulation of our problem of interest in Section~\ref{Ss:problem_formulation}.

\subsection{Graphs and graph neural networks}
\label{Ss:graphs_and_gnns}

Let $G = (V,E)$ be an undirected and unweighted graph where $V$ is the set of nodes with cardinality $|V|=n$, and $E$ is the set of edges such that the unordered pair $(i,j) \in E$ only if there exists an edge between nodes $i$ and $j$.
We adopt the convention where the neighborhood $N(u)$ of a node $u \in V$ includes itself and all the nodes connected to it , i.e., $N(u) = \{u\} \cup \{ v \in V | (u,v) \in E\}$.
Moreover, we consider the case where nodes in $V$ have associated features and labels.
Node features $\bbx_u \in \mathbb{R}^{d}$ represent node descriptors that can encode information that goes beyond the graph structure.
Also, every node $u$ may be associated with a label or class $y_u \in \mathcal{C}$, where $\mathcal{C}$ is a discrete set of labels.
We denote a (featured)\footnote{By default, all considered graphs will have features.} labeled graph as the tuple $(G, \bbX, \bby)$, where the rows of $\bbX \in \reals^{n \times d}$ correspond to the features of every node and $\bby \in \ccalC^n$ collects the node labels.
Analogously, a tuple $(G, \bbX)$ represents an unlabeled graph.

GNNs are a general class of neural architectures that leverage the underlying connectivity structure of a graph to learn suitable node-level and graph-level representations for various downstream tasks~\cite{wu2020comprehensive}. 
A generic GNN architecture $\Phi_G: \mathbb{R}^{n\times d} \to \mathbb{R}^{n\times o}$ is a learnable parametric transformation of the feature space of a graph $G$ through a sequence of layers.
We denote the output of a GNN as $\Phi_G(\bbX; \bbW)$, where $\bbW$ contains the parameters of the network.
Each layer of a GNN is composed of two basic operations: \emph{neighborhood aggregation} followed by \emph{feature transformation}. 
While a multitude of architectures~\cite{kipf2016semi,velivckovic2017graph,hamilton2017inductive,chen2018fastgcn} has been proposed, each offering distinct structures to one or more of these operations, the core principles for an arbitrary node $u$ at any layer $l$ of an $L$-layered GNN can be formalized as 
\begin{align}\label{eq:gcn}
    \bbz^{(l+1)}_u = \sigma \left( \frac{1}{\sqrt{d_l}} c_u \bbW^{(l)} \hspace{-0.1cm}  \sum_{v \in N(u)} \hspace{-0.1cm}\bbz^{(l)}_v  \right),
\end{align}
where $\bbz^{(0)}_u = \bbx_u$, $\bbW^{(l)} \in \mathbb{R}^{d_{l+1}\times d_{l}}$ contains trainable weights such that $\bbW = \{\bbW^{(0)}, \ldots,\bbW^{(L-1)}\}$, and $c_u$ is a normalizing factor for neighborhood density. 
We adopt the common choice of $c_u = 1/|N(u)|$.
Through these layered operations, the final output $\bbz^{(L)}_u \in \mathbb{R}^o$ fuses the local neighborhood information with the attributes of a given node $u$ to generate rich node-level representations.

\subsection{Neural tangent kernels}
\label{Ss:ntks}

Consider a general neural network $f (\bbx; \bbomega)$ where $\bbx$ is the input and $\bbomega$ is a vector of weights, which have been trained by minimizing a squared loss.
We can then approximate the learned function through a first-order Taylor expansion on the weights as
\begin{equation}\label{E:taylor_expansion}
   f (\bbx; \bbomega) \approx f (\bbx; \bbomega_{0}) + \nabla_{\bbomega} f (\bbx; \bbomega_{0})^\top (\bbomega - \bbomega_{0}),
\end{equation}
where $\bbomega_0$ denotes the weights at initialization.
In general, the approximation in~\eqref{E:taylor_expansion} would be extremely crude since the learned weights $\bbomega$ would differ significantly from $\bbomega_0$.
However, it has been observed in practice and theoretically shown that for severely \emph{over-parametrized} neural networks, the parameters $\bbomega$ only barely change during training~\cite{jacot2018neural}.
In such a setting, the approximation in~\eqref{E:taylor_expansion} is quite accurate and the neural network is close to a linear function of the weights $\bbomega$.
Hence, minimizing a squared loss reduces to just solving a linear regression.
Notice, however, that the approximation in~\eqref{E:taylor_expansion} is still non-linear in the input $\bbx$.
In fact, it is linear on a very specific feature transformation given by $\eta(\bbx) = \nabla_{\bbomega} f (\bbx; \bbomega_{0})$.
This feature transformation naturally induces a kernel on the input, which is termed the NTK~\cite{jacot2018neural}.
To be more precise, the NTK entry between two inputs $\bbx_i$ and $\bbx_j$ is given by $\nabla_{\bbomega} f (\bbx_i; \bbomega_{0})^\top \nabla_{\bbomega} f (\bbx_j; \bbomega_{0})$.
Reformulating a neural architecture as an NTK generates closed-form expressions to compute its output without performing explicit gradient updates on its parameters.
More recently, this idea was extended to GNNs to study their performance in the over-parametrized (infinite-width) limit~\cite{du2019graph,chen2019powerful,garg2020generalization}. 

\subsection{Problem formulation}
\label{Ss:problem_formulation}

Our goal is to learn from a set of fully-labeled graphs in order to estimate the labels of a different completely unlabeled graph.
Formally, we state our problem as follows.

\begin{problem}\label{P:main}
Given a set of $m$ labeled graphs $\{(G_i, \bbX_i, \bby_i)\}_{i=1}^m$, provide an estimate $\hat{\bby}_0$ of the labels of a given (unlabeled) graph $(G_0, \bbX_0)$.
\end{problem}

\noindent 
The above problem is of practical relevance when the labeled and unlabeled graphs are expected to share some common traits.
For example, we can have a series of social networks with features associated with each person (such as age, nationality, or salary) along with a label of interest such as whether they bought a given product or not.
Then, our unlabeled social network $(G_0, \bbX_0)$ can represent a new untapped market and our goal is to estimate the individuals that might be interested in the product.
As another example, consider the case where the labeled graphs represent brain networks from several individuals and the labels indicate whether the brain regions were significantly involved in performing a given task. Then, given a new individual, we want to predict which brain regions will become particularly active when the task is performed.

Both mentioned examples highlight the inductive nature of the problem at hand.
Unlike the (more common) problem of semi-supervised learning where we are given a partially labeled graph and we want to propagate those labels to the rest of the graph, in Problem~\ref{P:main} we have no connections between labeled and unlabeled nodes.
Instead, our task is to extract a fundamental way to relate the structure and features of a graph with its labels, and then apply these learned relations on a new graph $(G_0, \bbX_0)$ that is unavailable during training.

At a high level, we tackle Problem~\ref{P:main} by defining a \emph{similarity measure between nodes that potentially belong to different graphs}.
Having access to this similarity, we can rely on a kernel-based method to train a node classifier.
Then, given a new (unlabeled) graph, we can compute the similarities between its nodes and those in our training set, and apply the trained classifier to estimate its labels.
For the essential step of defining the aforementioned similarity, we rely on infinitely-wide GNNs with residual connections, as explained in Section~\ref{S:Solution}.

\section{Node Classification using Graph Neural Tangent Kernels}\label{S:Solution}

\begin{figure*}[t]
\centering
\includegraphics[width=\linewidth]{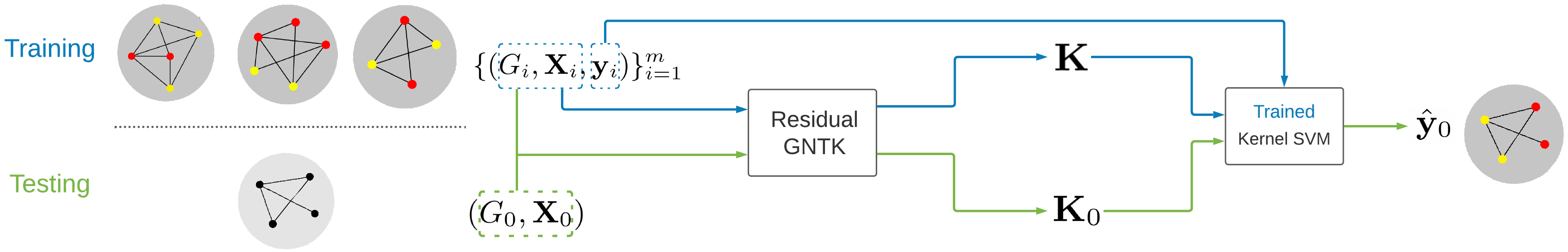}
\vspace{-5mm}
\caption{Scheme of our proposed pipeline. Following Proposition~\ref{pr:mainRes} we compute the GNTK with residual connections $\bbK$ between all the nodes in the training set.
We use $\bbK$ along with the given labels to train a kernel SVM classifier.
Given a new unlabeled graph $(G_0, \bbX_0)$ we first compute the GNTK entries between the nodes in the new graph and those in all the training graphs to obtain $\bbK_0$. Finally, we rely on our kernel SVM classifier to estimate the labels $\hat{\bby}_0$.}
\vspace{-3mm}
\label{fig:pipeline_scheme}
\end{figure*}

One could define a similarity measure between nodes based entirely on their features by, e.g., applying a decreasing function to any metric in the feature space.
In this way, two nodes that have feature vectors lying close to each other in feature space will have a high similarity value.
An immediate drawback of this approach is that it completely ignores the topological information of the graphs.
Alternatively, we could also define an entirely topological notion of similarity by, e.g., comparing the centrality values of the nodes.
In this way, two central nodes, even if belonging to different graphs, will be deemed similar to each other.

With the objective of finding a good solution to Problem~\ref{P:main}, we rely on GNTKs to define a similarity (kernel) between nodes that combines both sources of information (features and topology).

\subsection{GNTKs with residual connections}

The incorporation of residual or skip connections~\cite{he2016deep} in neural architectures is a general method for creating additional routes for forward propagation of hidden features and backward propagation of gradients. 
The main advantages of this technique include reduction in model complexity by conferring the model the flexibility needed to skip unnecessary layers and allowance for improved gradient flow to lower layers~\cite{veit2016residual}.  
In addition to other classes of connectionist architectures, residual connections have also been shown to significantly improve convergence behavior of GNNs through implicit acceleration~\cite{xu2021optimization}.
In this context, it is of interest to establish the expression of the GNTK that corresponds to an infinitely-wide GNN \emph{with residual connections}.

To be more precise, in a generic GNN architecture with $1$-skip connections, we replace the layer update in~\eqref{eq:gcn} with
\begin{align}\label{eq:res}
    \bbz^{(l+1)}_u = \sigma \Bigg(\frac{1}{\sqrt{d_l}} c_u \bbW^{(l)}_1 \hspace{-0.1cm} \sum_{v \in N(u)} \hspace{-0.1cm} \bbz^{(l)}_v +  \frac{1}{\sqrt{d_{l}}}\bbW^{(l)}_2\bbz^{(l)}_u \Bigg),
\end{align}
where we have now added a second term representing the residual connection with its own set of learnable parameters $\bbW^{(l)}_2 \in \mathbb{R}^{d_{l+1}\times d_{l}}$. 

As explained in Section~\ref{Ss:ntks}, our goal is to extract the kernel induced by a GNN's severely over-parametrized layers with the form in~\eqref{eq:res}.
In essence, given any two graphs $G=(V,E)$ and $G^{\prime}=(V^{\prime}, E^{\prime})$ with $n$ and $n^{\prime}$ nodes, respectively, we seek to recover a similarity matrix $\bbTheta \in \mathbb{R}^{n \times n^{\prime}}$ such that $[\bbTheta]_{u u^{\prime}}$ captures the similarity between nodes $u \in V$ and $u^{\prime} \in V^{\prime}$.
Following the now-established procedure for the computation of NTKs~\cite{jacot2018neural,arora2019exact}, we provide a recurrent formula for the computation of $\bbTheta$, where we propagate the similarities through the $L$ layers of our infinitely-wide GNN. 
We denote by $\bbTheta^{(l)}$ the similarity matrix after $l$ layers of our GNN and define $\bbTheta = \sum_{l=1}^L \bbTheta^{(l)}$. 
This modality where the similarities are aggregated across layers is sometimes referred to as the jumping knowledge variant~\cite{du2019graph}.
Alternatively, one can define $\bbTheta = \bbTheta^{(L)}$.
However, due to its better empirical performance, we adopt the jumping knowledge modality.
Moreover, the computation of $\bbTheta^{(l)}$ relies on another series of matrices $\bbSigma^{(l)}$ that are also recursively updated.
In their entries $\big[\bbSigma^{(l)} \big]_{uu^{\prime}}$, these matrices encode the covariance between Gaussian processes $\Psi_{l}(u)$ and $\Psi_{l}(u')$, which are instrumental for the derivation of the GNTK and arise when considering the infinitely wide setting, i.e., $d_1,\dots,d_{L-1} \rightarrow \infty$.
Having introduced this notation, the following result holds.

\begin{myproposition}\label{pr:mainRes}
Given two graphs $G$ and $G'$, the GNTK $\bbTheta = \sum_{l=1}^L \bbTheta^{(l)}$ of the residual GNN with layers as described in~\eqref{eq:res} is given (elementwise) by the following recursive procedure
\begin{align}
    \big[\bbTheta^{(l+1)}\big]_{uu^{\prime}} &= \big[\bbSigma^{(l+1)} \big]_{uu^{\prime}} + \big[\bbTheta^{(l)} \big]_{uu^{\prime}} \, \overline{\Psi}_l(u,u';\dot{\sigma}) \nonumber \\
&+ c_u c_{u^{\prime}} \hspace{-0.3cm} \sum_{v \in N(u)} \sum_{v^{\prime} \in N(u^{\prime})} \hspace{-0.2cm} \Big(\big[\bbTheta^{(l)} \big]_{vv^{\prime}} \overline{\Psi}_l(v,v';\dot{\sigma}) \Big), \label{E:gntk_main}
\end{align}
where 
\begin{align}
    \overline{\Psi}_l(u,u';\dot{\sigma}) &= \mathbb{E}_{\Psi_l \sim  \mathcal{N}(0,\bbSigma^{(l)})}\big[
\dot{\sigma}(\Psi_l(u))\dot{\sigma}(\Psi_l(u^{\prime}))
\big], \label{E:gntk_aux1} \\
\big[\bbTheta^{(1)} \big]_{uu^{\prime}} &= \big[\bbSigma^{(1)} \big]_{uu^{\prime}}, \label{E:gntk_aux2} \\
\big[\bbSigma^{(1)} \big]_{uu^{\prime}} &= \frac{1}{d} \bbx_u^{\top} \bbx_{u'} + \frac{1}{d}c_u c_{u^{\prime}} \!\!\!\! \sum_{\substack{v \in N(u) \\ v^{\prime} \in N(u^{\prime})}} \!\!\! \bbx_v^\top \bbx_{v'}, \label{E:gntk_aux3} \\
\big[\bbSigma^{(l+1)} \big]_{uu^{\prime}} &= \overline{\Psi}_l(u,u';{\sigma})  +  c_u c_{u^{\prime}} \sum_{\substack{v \in N(u) \\ v^{\prime} \in N(u^{\prime})}} \overline{\Psi}_l(v,v';{\sigma}), \label{E:gntk_aux4}
\end{align}
and $\dot{\sigma}$ denotes the derivative of $\sigma$.
\end{myproposition}
\begin{myproof}
The above result can be proven by induction similar to the proof derived in~\cite[Prop. 1]{jacot2018neural} for multi-layer perceptrons (MLPs). Essentially, we can consider the propagation rule in~\eqref{eq:res} as a combination of an MLP given by $\bbW^{(l)}_2$ and the aggregation term corresponding to a standard GNN. The key difference between our approach and that of~\cite{du2019graph} is that by considering the residual layer as an MLP, the expression for the covariance $\big[\bbSigma^{(l+1)} \big]_{uu^{\prime}}$ in~\cite{du2019graph} is modified as in~\eqref{E:gntk_aux4}. 
Once that relation has been established, we can follow analogous steps to the derivation of $\big[\bbTheta^{(l+1)} \big]_{uu^{\prime}}$ in~\cite{jacot2018neural} to obtain the expression in Proposition~\ref{pr:mainRes}.
\end{myproof}

\begin{figure*}[t]
	\centering
	\subfigure[]{
			\centering
			\includegraphics[width=0.235\textwidth,height=0.18\textwidth]{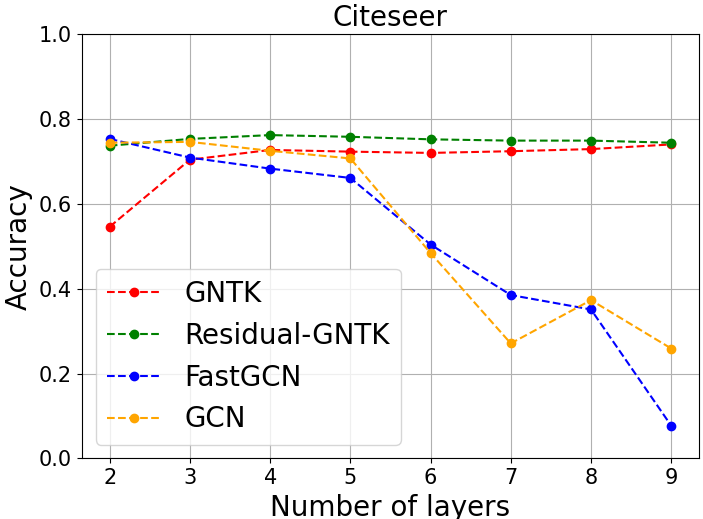}
			\label{Fig:cite}
		}	
	\subfigure[]{
			\centering
			\includegraphics[width=0.235\textwidth,height=0.18\textwidth]{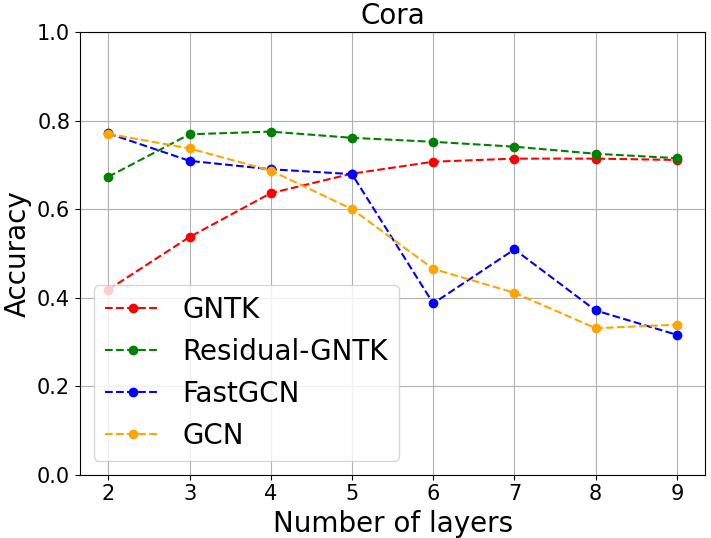}
			\label{Fig:cora}
		}	
	\subfigure[]{
			\centering
			\includegraphics[width=0.235\textwidth,height=0.18\textwidth]{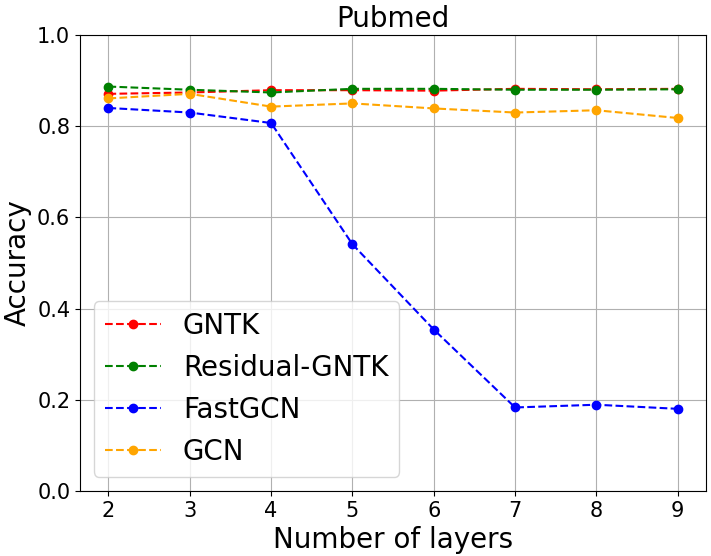}
			\label{Fig:pub}
		}	
	\subfigure[]{
			\centering
			\includegraphics[width=0.235\textwidth,height=0.18\textwidth]{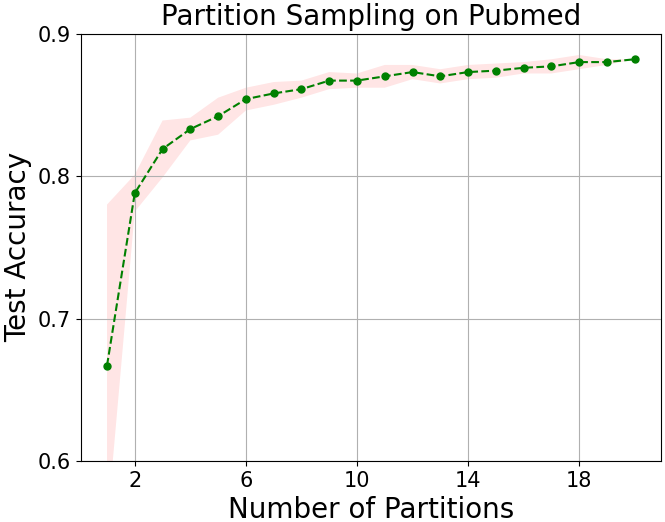}
			\label{Fig:sampl}
		}
		\vspace{-3mm}
		\caption{\small {Node classification on citation networks. (a-c)~Performance comparison in terms of test accuracy of GCN, FastGCN, vanilla GNTK, and Residual-GNTK on three benchmarked datasets: (a)~Citeseer, (b)~Cora, and (c)~Pubmed.
		(d)~Test accuracy of Residual-GNTK on Pubmed as a function of the number of randomly selected training graphs. }}
		\vspace{-3mm}
		\label{Fig:combined}
\end{figure*}

The key component of the above result is its treatment of the residual connection as an MLP that is added to to the aggregation structure of a GNN. 
This allows us to establish an explicit formula to capture the layer-wise evolution of the residual GNTK. 
A node-level similarity structure of this form allows us to learn suitable node representations based on corresponding node features of multiple graphs. 

\subsection{Label propagation across graphs}

Having established a relevant similarity measure between nodes that potentially reside on different graphs (cf. Proposition~\ref{pr:mainRes}), we now detail our node classification method by subsequently going over its training and testing phases.

\vspace{1mm}
\noindent {\bf Training.}
Given our training graphs $\{(G_i, \bbX_i, \bby_i)\}_{i=1}^m$ we compute the residual GNTK matrices $\bbTheta (G_i, G_j)$ following \eqref{E:gntk_main}-\eqref{E:gntk_aux4} for every pair of graphs, where we now explicitly state in the GNTK notation which graphs are involved in its computation.
We can then collect these $\bbTheta (G_i, G_j)$ into a kernel matrix ${\bbK} \in \reals^{\sum_{i=1}^m n_i \times \sum_{i=1}^m n_i}$, given as follows
\begin{equation}
   {\bbK} = 
   \begin{bmatrix}
   \bbTheta (G_1, G_1) & \cdots & \bbTheta (G_1, G_m) \\
   \vdots & \ddots & \vdots \\
   \bbTheta (G_m, G_1) & \cdots & \bbTheta (G_m, G_m)
   \end{bmatrix}.
\end{equation}
Notice that ${\bbK}$ stores a similarity measure between every pair of nodes in our training set.
Leveraging the fact that we have labels $\{\bby_i\}_{i=1}^m$ for all these nodes, we can train any kernel-based classifier on these data such as a kernel SVM or a kernel logistic regression~\cite{zhu2002support}.
Due to their widespread adoption, we select kernel SVMs in our implementation.
We do not provide a detailed description of how kernel SVMs work since this is not the focus of the paper. 
However, we direct the interested reader to~\cite{james2013introduction}.
The described training procedure is depicted in Fig.~\ref{fig:pipeline_scheme} using blue arrows.

\vspace{1mm}
\noindent {\bf Testing.}
When presented with the new unlabeled graph $(G_0, \bbX_0)$, we can compute the GNTK similarities between every node in $G_0$ and all the nodes in our training dataset and store them in ${\bbK}_0 \in \reals^{n_0 \times \sum_{i=1}^m n_i}$ as
\begin{equation}
   {\bbK}_0 = 
   \begin{bmatrix}
   \bbTheta (G_0, G_1) & \cdots & \bbTheta (G_0, G_m)
   \end{bmatrix}.
\end{equation}
Due to the nature of kernel methods~\cite{hofmann2008kernel}, the information in ${\bbK}_0$ is enough to obtain label estimates $\hat{\bby}_0$ from our previously trained kernel SVM. 
This testing pipeline is represented using green arrows in Fig.~\ref{fig:pipeline_scheme} and we can see its implementation in the next section.

\section{Numerical experiments}\label{S:num_exp}

We empirically demonstrate the performance of our GNTK with residual connections (Residual-GNTK) for node classification on citation networks. 
For our experiments, we consider a purely inductive setting -- test graph is completely isolated from the training graph. 
A description of the datasets is provided next, followed by a comparison of our model performance with that of GCN~\cite{kipf2016semi} and FastGCN~\cite{chen2018fastgcn}, in terms of classification accuracy. 
Finally we discuss the effect of partitioning a large training graph into multiple sub-graphs of amenable sizes and using only a subset of these smaller graphs for training a classifier.

\myparagraph{\textbf{Datasets}} We use three benchmarked public citation networks of different sizes: Cora, Citeseer, and Pubmed~\cite{kipf2016semi}. 
While Cora and Citeseer are of amenable size ($<5$k nodes), Pubmed has about $19$k nodes, thus, we rely on METIS~\cite{karypis1998fast} to partition this graph into smaller subgraphs.
In particular, we obtain $20$ subgraphs of equal size following the data splits in FastGCN~\cite{chen2018fastgcn} with the modification that we remove all the edges connecting the test graphs to the training and validation graphs for each dataset to perfectly emulate an inductive setting.

\myparagraph{\textbf{Performance comparison}} We compare the node classification accuracy achieved by Residual-GNTK with that of vanilla GNTK and two existing solution models, FastGCN~\cite{chen2018fastgcn} and GCN~\cite{kipf2016semi}. 
For GNTK and Residual-GNTK, we train a support vector machine classifier (SVM-C) with $\bbK$ as the precomputed kernel, and test on the completely unseen graph using $\bbK_0$. 
The comparisons are shown in Fig~\ref{Fig:combined}. 
While there is a clear dip in performance of GCN and FastGCN for all three datasets as evident in Fig~\ref{Fig:cite},~\ref{Fig:cora} and~\ref{Fig:pub} indicating oversmoothing, the GNTK models maintain a consistent performance even for deeper architectures. 
Moreover, Residual-GNTK shows a consistent improvement in accuracy over that of vanilla GNTK, clearly translating the effectiveness of residual connections to the kernel formulation.

\myparagraph{\textbf{Scalability}} We now explore the possibility of using a small subset of partitions to train Residual-GNTK, especially for large graphs. To this end, we randomly select $m \in \{1, 2, \ldots, 20\}$ partitions of the training graph of Pubmed to train an SVM-C using a Residual-GNTK of $5$ layers. We present the mean and standard deviation of $10$ trials for each $m$ in Fig~\ref{Fig:sampl}.
We can observe a $10$-fold decrease in the standard deviation of classification accuracy when $2$ partitions are used instead of $1$. 
Furthermore, the steep increase of accuracy as we move from $1$ to $10$ sub-graphs suggests that it is possible to achieve sufficiently high accuracy even by using just about half of the training graphs, sampled randomly.

\section{Conclusions}\label{S:Conclusions}

We have presented a method to estimate all the labels of an entirely unlabeled graph. The method relies on calculating similarities between the unlabeled nodes and the labeled training nodes belonging to different graphs. The similarities are computed based on a proposed variation of the GNTK, which are in turn fed to a kernel-based classifier.
Future research avenues include: i)~The derivation and implementation of other GNTKs based on variants of the underlying GNN, and ii)~The design of scalable and parallelizable pipelines that rely on ensemble classifiers for cases where the number of training graphs $m$ is large.

\newpage
\bibliographystyle{IEEEbib}
\bibliography{strings,refs}

\end{document}

%% file: my_sections.tex
\usepackage{needspace}

\newcommand{\myparagraph}[1]{\needspace{1\baselineskip}\medskip\noindent {\it #1.}}



%% file: figures/tikz_styles.tex
\tikzstyle{phantom vertex} = [ ellipse, 
                               anchor = center, 
                               minimum height = 1*\unit, 
                               minimum width  = 1*\unit,
                               inner sep=0pt,
                               anchor=center]
\tikzstyle{red vertex}   = [black, fill = red!20,   phantom vertex, draw]
\tikzstyle{black vertex} = [black, fill = black!20, phantom vertex, draw]
\tikzstyle{blue vertex}  = [black, fill = blue!20,  phantom vertex, draw]
\tikzstyle{green vertex} = [black, fill = green!20,  phantom vertex, draw]
\tikzstyle{yellow vertex} = [black, fill = yellow!20,  phantom vertex, draw]
\tikzstyle{cyan vertex} = [black, fill = cyan!20,  phantom vertex, draw]
\tikzstyle{vertex}       = [draw, phantom vertex]

\tikzstyle{point} = [ellipse, inner sep=0pt, draw, fill=white, anchor = center,
                     minimum height = 0.05*\unit, minimum width  = 0.05*\unit]